\documentclass[10pt,twocolumn,letterpaper]{article}

\usepackage{iccv}              %

\usepackage{tabularx, booktabs, adjustbox}
\usepackage{makecell}
\usepackage{graphicx}
\usepackage{subcaption}
\usepackage{amsmath}
\usepackage{amssymb}
\usepackage{amsfonts}
\usepackage{bm} 
\usepackage{siunitx}
\usepackage{pifont}
\usepackage{soul}
\usepackage[normalem]{ulem}

\definecolor{iccvblue}{rgb}{0.21,0.49,0.74}
\usepackage[pagebackref,breaklinks,colorlinks,allcolors=iccvblue]{hyperref}

\def\paperID{*****} %
\def\confName{ICCV}
\def\confYear{2025}

\title{AssemblyHands-X: Modeling 3D Hand-Body Coordination\\for Understanding Bimanual Human Activities}
\author{Tatsuro Banno \quad Takehiko Ohkawa \quad Ruicong Liu \quad 
Ryosuke Furuta \quad 
Yoichi Sato \\
The University of Tokyo, Tokyo, Japan\\
{\tt\small \{banno, ohkawa-t, lruicong, furuta, ysato\}@iis.u-tokyo.ac.jp}
}

\begin{document}
\maketitle
\begin{abstract}

Bimanual human activities inherently involve coordinated movements of both hands and body.
However, the impact of this coordination in activity understanding has not been systematically evaluated 
due to the lack of suitable datasets.
Such evaluation demands kinematic-level annotations (\eg, 3D pose) for the hands and body, yet existing 3D activity datasets typically annotate either hand or body pose.
Another line of work employs marker-based motion capture to provide full-body pose, but the physical markers introduce visual artifacts, thereby limiting models' generalization to natural, markerless videos.
To address these limitations, we present \emph{AssemblyHands-X}, the first markerless 3D hand-body benchmark for bimanual activities, designed to study the effect of hand-body coordination for action recognition.
We begin by constructing a pipeline for 3D pose annotation from synchronized multi-view videos.
Our approach combines multi-view triangulation with SMPL-X mesh fitting, yielding reliable 3D registration of hands and upper body.
We then validate different input representations (\eg, video, hand pose, body pose, or hand-body pose) across recent action recognition models based on graph convolution or spatio-temporal attention.
Our extensive experiments show that pose-based action inference is more efficient and accurate than video baselines.
Moreover, joint modeling of hand and body cues improves action recognition over using hands or upper body alone, highlighting the importance of modeling interdependent hand-body dynamics for a holistic understanding of bimanual activities.

\end{abstract}

\section{Introduction}
\begin{figure}[t]
    \centering
    \includegraphics[width=\linewidth]{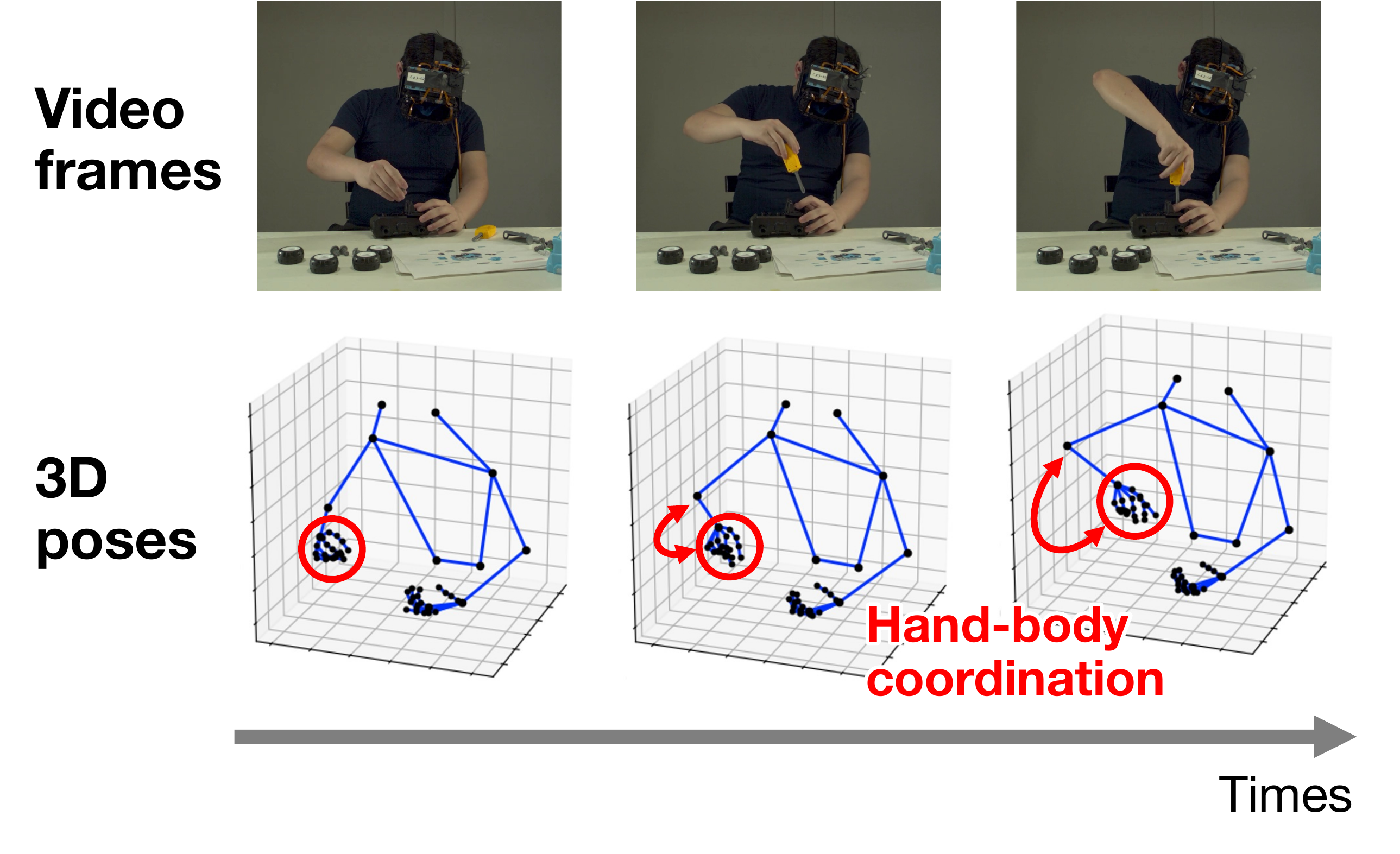}
    \caption{\textbf{Bimanual human activities inherently involve coordinated movements of both hands and body.} Our proposed benchmark, AssemblyHands-X, provides the first markerless 3D hand-body dataset,
    enabling systematic evaluation for the effect of hand-body coordination in action recognition.}
    \label{fig:teaser}
\end{figure}

When we imagine nuanced everyday activities like screwing with a screwdriver, we intuitively grasp the deep coordination between our hands and body. 
As shown in Fig.~\ref{fig:teaser}, precise hand movements coincide with forearm rotation and elbow bends. 
This tight hand-body coupling forms the distinct motion patterns of bimanual activities. Observing this connection brings us a question: \emph{how can modeling these interdependent hand-body dynamics improve the understanding of bimanual activities?}

\begin{figure*}[t]
	\begin{center}
		\includegraphics[width=0.97\linewidth]{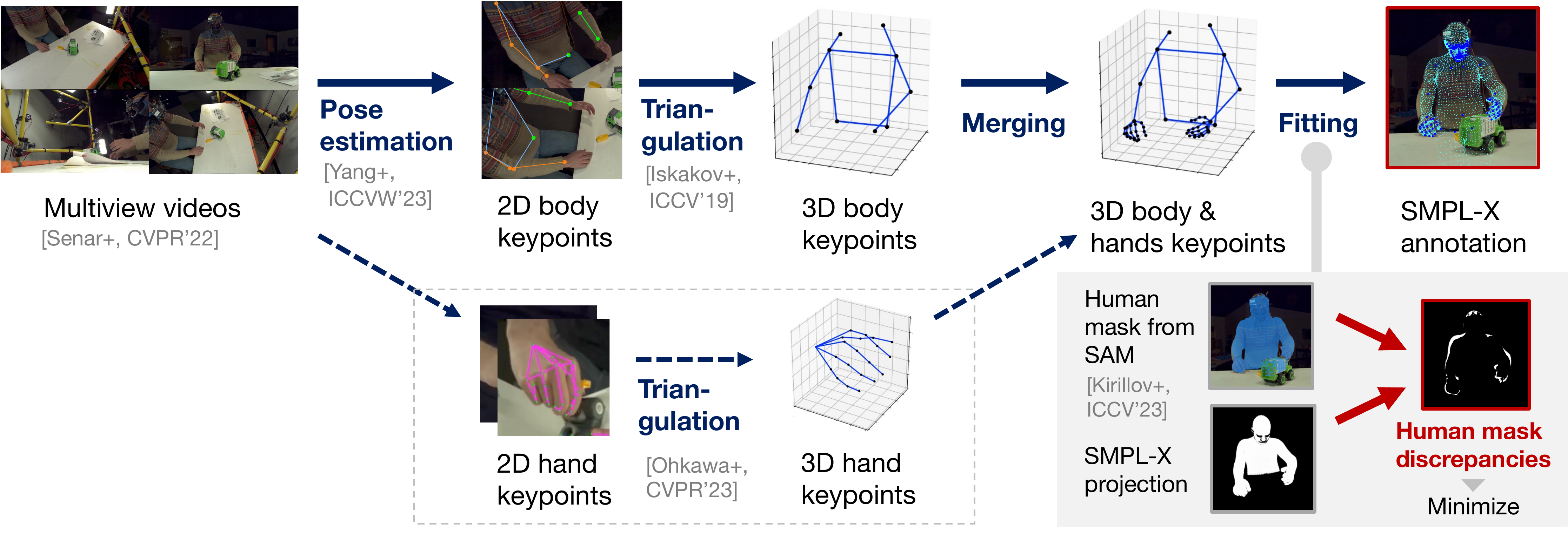}
		\caption{\textbf{Dataset annotation pipeline of AssemblyHands-X.} We first crop the body and hand regions, perform independent multi-view triangulation, and merge the results to obtain 3D hand-body keypoints. These keypoints are then used to fit the SMPL-X model, producing a unified 3D representation of human motion.}
		\label{fig:pipeline}
	\end{center}
\end{figure*}

However, answering the above question is difficult due to the lack of suitable benchmarks that provide kinematic-level annotations (\eg, 3D pose) for hand-body dynamics. 
Most existing benchmarks for bimanual activities~\cite{rai2021home, zheng2023ha} focus on video-based action recognition~\cite{lin2019tsm}, lacking explicit 3D hand-body signals. 
While recent activity datasets provide 3D body pose~\cite{ben2021ikea, aganian2023attach}, they struggle to obtain detailed hand information.
Due to the complex articulation of hands, they typically omit annotating hand pose~\cite{ben2021ikea} or provide limited hand signals (\eg, three keypoints in~\cite{aganian2023attach}).
Conversely, 
prior works on fine-grained hand interactions capture rich 3D hand poses from egocentric~\cite{sener2022assembly101} or multi-view~\cite{ohkawa2023assemblyhands, fan2024benchmarks} setups, while they typically neglect capturing body pose due to their tailored camera setups for close-up hand areas.
Furthermore, while recent marker-based motion capture 
can record precise 3D hand-body motion~\cite{fan2023arctic, zhan2024oakink2}, the use of physical markers introduces inevitable visual artifacts, 
resulting in models with limited generalization to natural, markerless videos.

To address these limitations, we introduce \emph{AssemblyHands-X}, the first markerless 3D hand-body benchmark for bimanual activities, designed for systematic evaluation for the effect of hand-body coordination on action recognition. Building on multi-view videos from Assembly101~\cite{sener2022assembly101}, we develop an automatic annotation pipeline of 3D poses, which combines multi-view triangulation with SMPL-X~\cite{pavlakos2019expressive} fitting. 
This enables the reliable recovery of synchronized 3D hand and body poses without relying on maker-based motion capture systems~\cite{fan2023arctic, zhan2024oakink2}.

Based on our new hand-body annotations, we validate different input representations for action recognition (\eg, video, hand pose, body pose, or hand-body pose) on recent backbones using graph convolutional~\cite{liu2020disentangling, zhou2024blockgcn} and spatio-temporal attention~\cite{wu2024frequency}. 
Our extensive experiments show that pose-based action inference is both more efficient and accurate than video-based baselines.
This shows the effectiveness of using 3D pose as a compact and discriminative representation for action recognition. 
Furthermore, joint modeling of hands and body
consistently improves action recognition compared to using either modality alone, highlighting that capturing interdependent hand-body dynamics benefits a holistic understanding of bimanual activities.

The main contributions of this work are as follows:

\begin{itemize}
\item We introduce AssemblyHands-X, the first markerless 3D hand-body benchmark for bimanual activities, designed for the systematic evaluation of the effect of hand-body coordination in action recognition.
\item We introduce a multi-view annotation pipeline with triangulation and SMPL-X model fitting, which enables reliable capture of 3D poses for both hands and body without relying on maker-based motion capture systems.
\item Using AssemblyHands-X, we evaluate different input representations for action recognition (\eg, video, hand pose, body pose, or hand-body pose), demonstrating that modeling interdependent hand-body dynamics is crucial for a holistic understanding of bimanual activities.

\end{itemize}

\section{Dataset annotation pipeline}

\subsection{Overview}
An overview of our AssemblyHands-X dataset annotation pipeline is illustrated in Fig.~\ref{fig:pipeline}. The input consists of multi-view videos from Assembly101~\cite{sener2022assembly101}, and the output includes 3D keypoints for the hands and upper body, along with corresponding parameters of a parametric whole-body human model (\ie, SMPL-X~\cite{pavlakos2019expressive}).

The pipeline begins by separately cropping the body and hands and performing multi-view triangulation independently for each. For hand pose estimation, we adopt the precise triangulation pipeline proposed by Ohkawa \etal~\cite{ohkawa2023assemblyhands}. In contrast, accurate body pose estimation during bimanual activities presents unique challenges, such as body truncation due to limited camera viewpoints. To address these issues, we propose a new body-specific triangulation pipeline (Sec.~\ref{label:body}). We then combine the estimated 3D body and hand joints to generate coordinated 3D hand-body keypoint annotations. Finally, leveraging these 3D keypoints along with human segmentation masks extracted from the multi-view images~\cite{kirillov2023segment}, we fit the SMPL-X parametric model to each frame (Sec.~\ref{label:smplx}), providing a kinematically consistent 3D representation of human motion.

The following sections provide a detailed description of each stage of the annotation pipeline, with particular emphasis on the body triangulation pipeline and the SMPL-X fitting process.

\subsection{Body triangulation pipeline}
\label{label:body}
\noindent{\textbf{2D keypoint detection.}}
For each cropped frame, we first apply a 2D keypoint estimation model~\cite{yang2023effective} for each camera view $c = 1, 2, \dots, C$, obtaining 2D locations $(x_c, y_c) \in \mathbb{R}^2$ and confidence scores $w_c \in [0, 1]$ for each body joint. We then apply median filtering to each keypoint across frames in every view to reduce short-term fluctuations.

Next, to suppress overconfident 2D keypoints near crop boundaries, we modulate each keypoint’s confidence $w_c$ based on its distance to the image edges. The adjusted confidence $w'_c$ is computed as:
\begin{equation}
w'_c = 
\min\left( \frac{x_c}{m}, \frac{W - x_c}{m}, \frac{y_c}{m}, \frac{H - y_c}{m}, 1 \right)
\cdot w_c
\end{equation}
Here, $(x_c, y_c)$ is the keypoint location within a cropped size $W\times H$, and $m$ is a predefined margin width. This formulation ensures that keypoints farther than $m$ pixels from the crop edges retain their original confidence, while those within the margin are linearly downweighted as they approach the boundary. We then threshold the adjusted confidence scores at $\tau = 0.15$, setting all values below $\tau$ to zero and linearly rescaling values above $\tau$ to the range $[0,1]$. This effectively discards low-confidence keypoints while preserving and normalizing reliable ones for subsequent processing.\\

\noindent{\textbf{3D keypoint triangulation.}}
Next, we estimate 3D body joint locations by combining 2D keypoints from multiple views using a confidence-weighted linear triangulation~\cite{iskakov2019learnable}. 
Let the 2D position of a joint in camera $c = 1, \dots, C$ be $(x_c, y_c) \in \mathbb{R}^2$, with adjusted confidence score $w'_c \in [0, 1]$. Following~\cite{iskakov2019learnable}, we construct a matrix $A_j \in \mathbb{R}^{2C \times 4}$ from the component of camera projection matrices of each view and 2D joint locations (see~\cite{hartley2003multiple} for full details), and solve for the homogeneous 3D coordinate $\bm{X}_j \in \mathbb{R}^4$ using the following confidence-weighted least-squares equation:

\begin{equation}
    (\bm{w}' \circ A) \bm{X} = 0 
\end{equation}
Here, $\bm{w}' = (w'_1, w'_1, w'_2, w'_2, \cdots, w'_C, w'_C)^\top$, and $\circ$ denotes the Hadamard product, meaning that the $i$-th row of matrix $A$ is multiplied by the $i$-th element of the vector $\bm{w}'$.

This process yields the 3D locations of the body joints. By combining these with the precise 3D hand joint annotations from Ohkawa \etal~\cite{ohkawa2023assemblyhands}, we generate a unified set of 3D keypoints covering the entire upper body and hands.

\subsection{SMPL-X model fitting}
\label{label:smplx}

\begin{figure}[t]
	\begin{center}
		\includegraphics[width=0.85\linewidth]{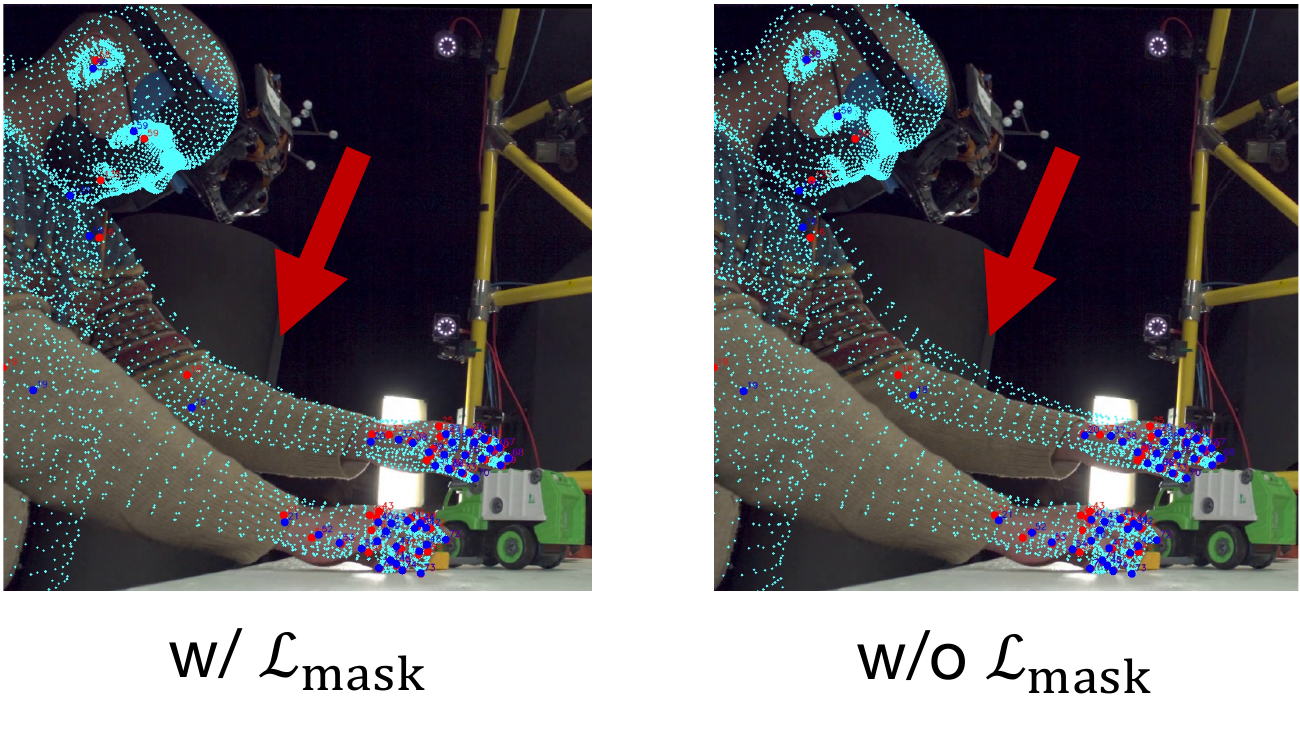}
        \captionsetup{skip=0.05pt}
		\caption{\textbf{Qualitative results of the fitting process with and without the silhouette loss \(\mathcal{L}_{\mathrm{mask}}\).} Incorporating \(\mathcal{L}_{\mathrm{mask}}\) significantly improves the alignment between the projected SMPL-X mesh and the observed human silhouette.
}
\vspace{-11pt}
		\label{fig:mask_vis}
	\end{center}
\end{figure}

Next, given the estimated 3D joint positions for both the hands and body, we optimize the SMPL-X parameters, namely the shape \(\bm{\beta}\) and the pose \(\bm{\theta}\), for each frame by minimizing the following objective:

\vspace{-15pt}
\begin{equation}
\mathcal{L} =  \lambda_{\mathrm{joint}} \mathcal{L}_{\mathrm{joint}} + \lambda_{\mathrm{mask}} \mathcal{L}_{\mathrm{mask}} + \lambda_{\beta}\|\bm{\beta}\|^2_2 + \lambda_{\theta} \|\bm{\theta}\|^2_2, 
\end{equation}
where each term is weighted by its corresponding coefficient \(\lambda\). The first term, \(\mathcal{L}_{\mathrm{joint}}\), penalizes the \(L_2\) distance between predicted and target 3D joint positions. The second term, \(\mathcal{L}_{\mathrm{mask}}\), is a silhouette loss~\cite{liu2019soft} that aligns the SMPL-X mesh projection with human segmentation masks extracted from the multi-view images~\cite{kirillov2023segment}, significantly improving the match between the projected mesh and the observed human silhouette (see Fig.~\ref{fig:mask_vis}).

\section{Experiments on action recognition}
\subsection{Experiment settings}
Using AssemblyHands-X, we conduct a comprehensive benchmark of action recognition. As input, we utilize 3D keypoints obtained prior to SMPL-X fitting and the corresponding video frames from \emph{View 3} of the Assembly101 ~\cite{sener2022assembly101}, which provides a clear view of both the body and hands. For the action labels, we adopt the six verb classes defined in Assembly101, following prior work on 3D hand pose-based action recognition~\cite{ohkawa2023assemblyhands}: \emph{pick up}, \emph{put down}, \emph{position}, \emph{remove}, \emph{screw}, and \emph{unscrew}. All pose sequences are standardized to a fixed length of $T=100$ frames, with shorter sequences padded by repeating the sequence.

\begin{figure*}[t]
    \centering

    \begin{subfigure}[t]{0.30\textwidth}
        \centering
        \includegraphics[width=\linewidth]{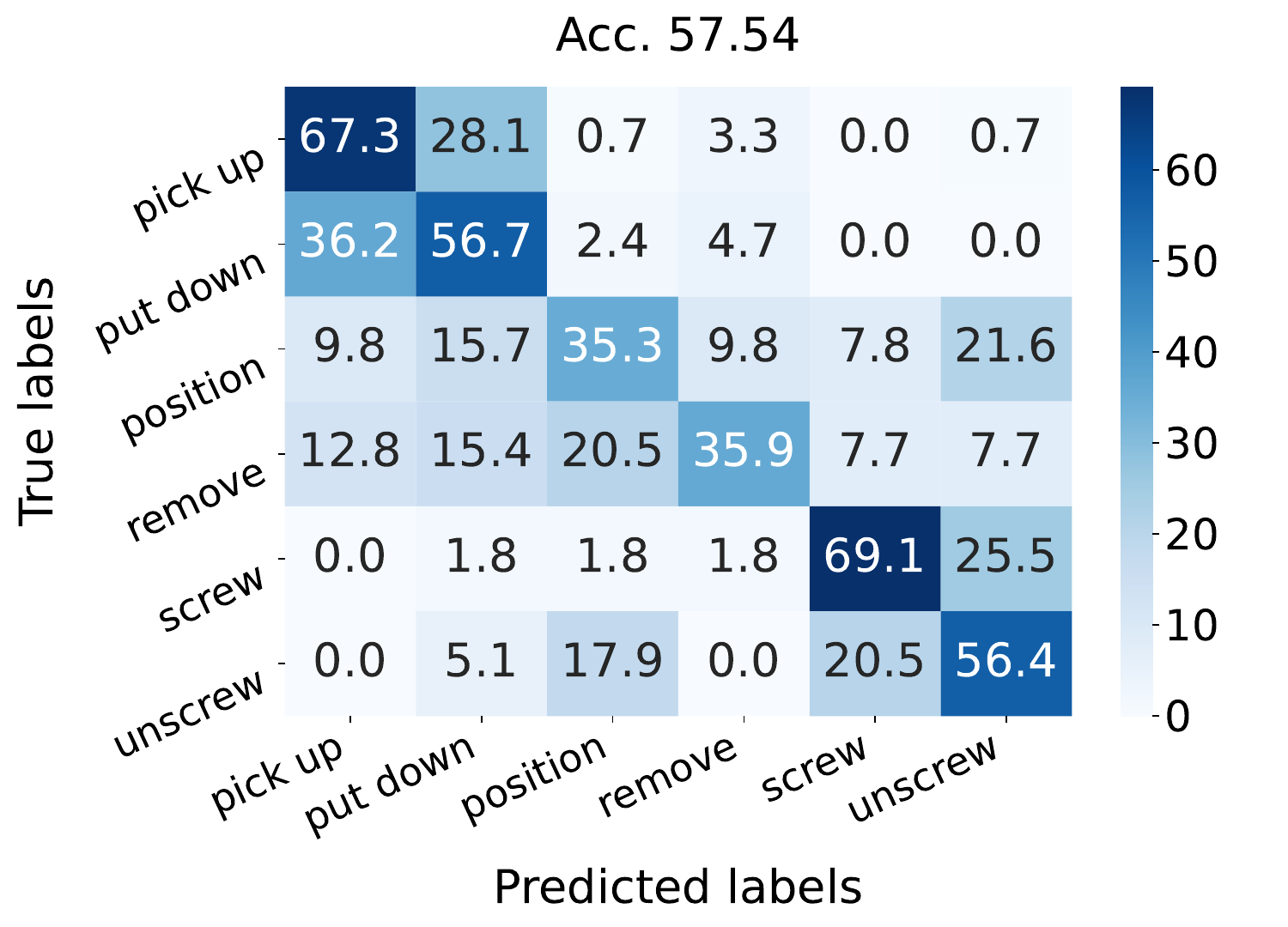}
        \caption{Body pose}
    \end{subfigure}
    \hfill
    \begin{subfigure}[t]{0.30\textwidth}
        \centering
        \includegraphics[width=\linewidth]{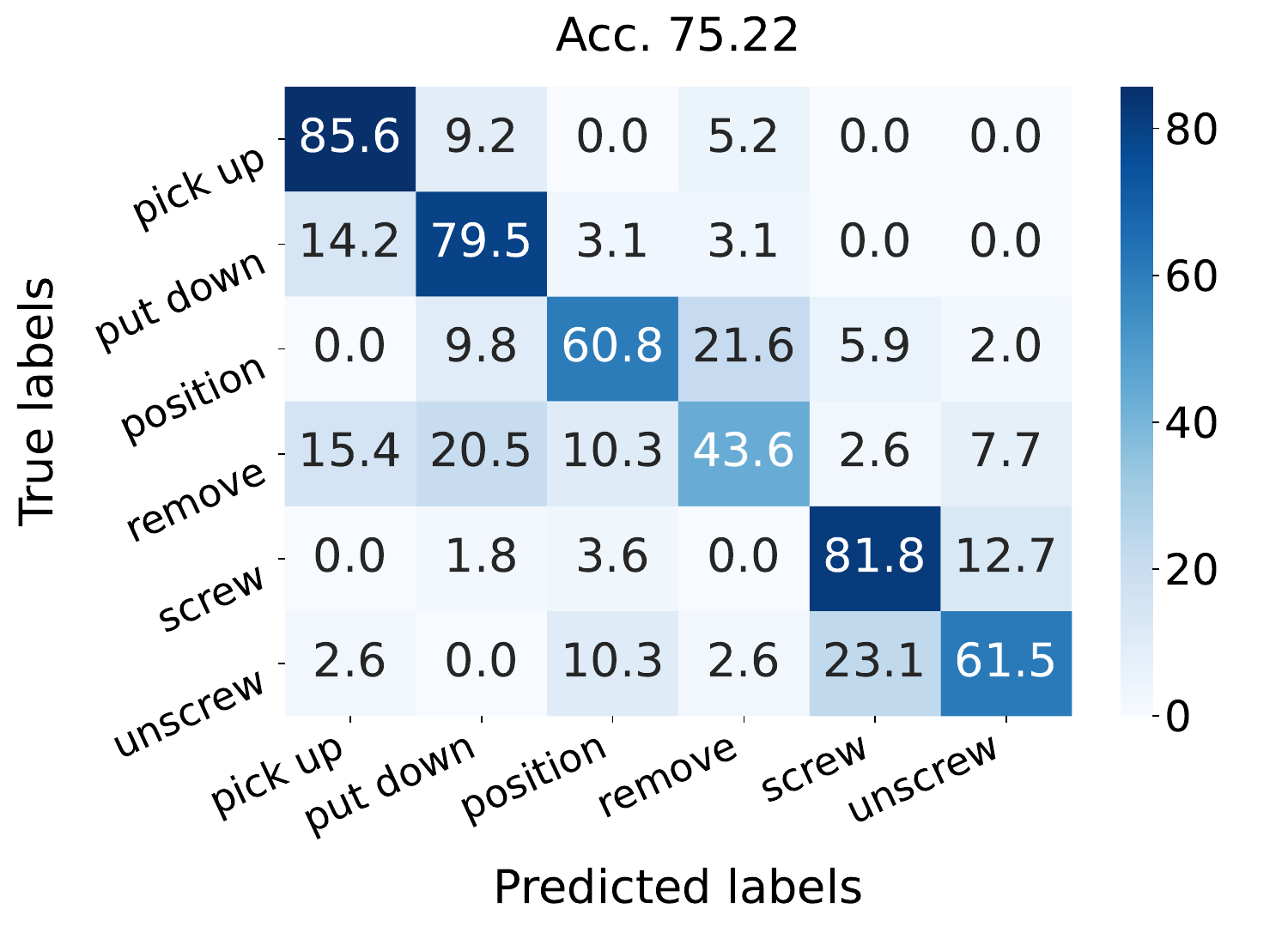}
        \caption{Hand pose}
    \end{subfigure}
    \hfill
    \begin{subfigure}[t]{0.30\textwidth}
        \centering
        \includegraphics[width=\linewidth]{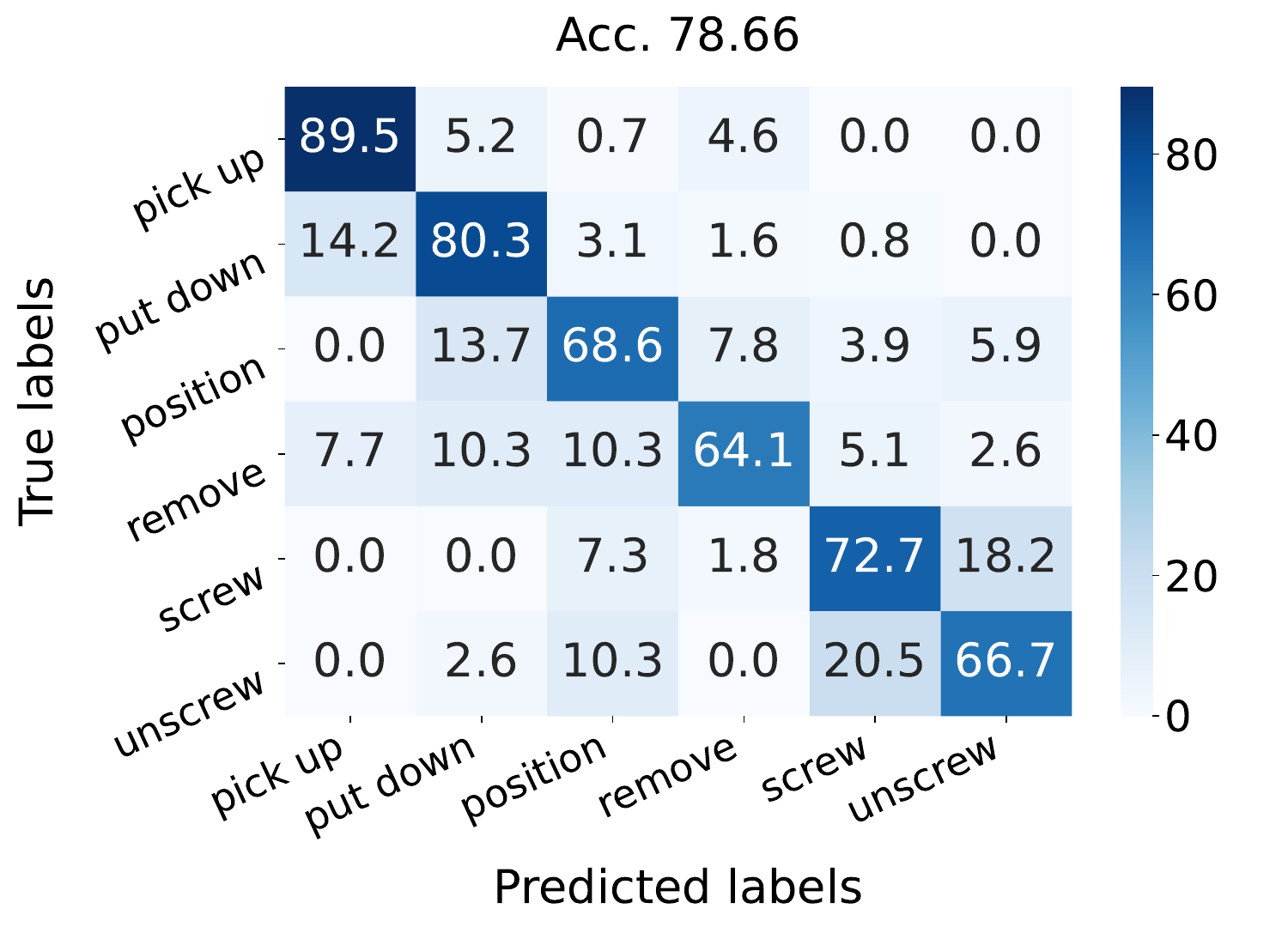}
        \caption{Combined hand-body pose}
    \end{subfigure}

    \caption{\textbf{Confusion matrices for action recognition under different 3D pose input configurations.} Each matrix corresponds to pose-based inference using FreqMixFormer~\cite{wu2024frequency}, trained and evaluated on the respective 3D pose configuration.}
    \label{fig:matrix}
\end{figure*}
\subsection{Effect of input representations}

We first evaluate different types of inputs on action recognition, including video, hand pose, body pose, and combined hand-body pose. For video-based inference, we consider two baselines: a large multimodal model (Gemini 2.5 pro~\cite{comanici2025gemini}) with strong zero-shot capability of video comprehension
and a standard video recognition model trained on our data (TSM~\cite{lin2019tsm}). For pose-based inference, we train graph convolutional networks (MS-G3D~\cite{liu2020disentangling}, BlockGCN~\cite{zhou2024blockgcn}) and 
a transformer baseline with spatio-temporal attention (FreqMixFormer~\cite{wu2024frequency}).\\

\noindent\textbf{Results.}
\begin{table}[t]
    \centering
    \caption{\textbf{Comparison of action recognition result under different input representations.} We report GFLOPs and Accuracy (\%) of each model on the AssemblyHands-X benchmark. The dagger (\(\dag\)) indicates results obtained in a zero-shot evaluation setting.}
    \label{tab:modal}
\begin{adjustbox}{width=0.40\textwidth}
\begin{tabular}{@{}llllll@{}} \toprule
& \multicolumn{2}{c}{3D pose} & &    \\ \cmidrule(lr){2-3} 
Method & Body & Hand & Video & GFLOPs &Acc. \\ \midrule
Random choice & - & -  & - & - & 16.7 \\
 Gemini 2.5 Pro\dag~\cite{comanici2025gemini} & \ding{55} & \ding{55}  & \ding{51}& - & 30.9 \\
TSM~\cite{lin2019tsm} & \ding{55} & \ding{55}  & \ding{51}& 32.88 & 62.3 \\\midrule
MS-G3D~\cite{liu2020disentangling}&\ding{51} & \ding{55}  & \ding{55}& 0.79 & 53.0 \\
& \ding{55} & \ding{51} & \ding{55}& 8.24  & 63.4 \\
&\ding{51} & \ding{51}  & \ding{55}& 10.16   & 69.6 \\ \midrule
BlockGCN~\cite{zhou2024blockgcn}&\ding{51} & \ding{55}  & \ding{55}& 0.34  & 54.1 \\
& \ding{55} & \ding{51} & \ding{55}& 2.89  & 72.6 \\
&\ding{51} & \ding{51}  & \ding{55}& 3.23   & 73.9 \\ \midrule
FreqMixFormer~\cite{wu2024frequency}&\ding{51} & \ding{55} & \ding{55}& 0.46   &57.5 \\
&\ding{55} & \ding{51}  & \ding{55}& 3.63   & \underline{75.2} \\
&\ding{51} & \ding{51}  & \ding{55}& 4.03   & \bfseries 78.7 \\\bottomrule
\end{tabular}
\end{adjustbox}
\end{table}
Table~\ref{tab:modal} shows action recognition result under different input representations. 
When comparing the best video-based model (TSM) and the best pose-based model (FreqMixFormer with 3D body-hand pose), the pose-based model achieves 16.4\% higher accuracy while requiring 8.2$\times$ fewer GFLOPs.  
This demonstrates that 3D pose, which explicitly captures fine-grained hand-body kinematics, provides a compact and discriminative representation for recognizing actions in bimanual activities.

Table~\ref{tab:modal} also shows that, across all pose-based models, the full body-hand representation consistently outperforms either pose alone. 
This suggests that integrating full body-hand information contributes to more accurate recognition of bimanual activities.
The corresponding confusion matrices for each 3D pose configuration (Fig.~\ref{fig:matrix}) further illustrate this effect, particularly showing that the hand-body model achieves substantial improvements on actions that are challenging for the hand-only model (\eg, \emph{position} and \emph{remove}).

\subsection{Effect of 3D pose annotation quality}
\begin{table}[t]
    \centering
    \caption{\textbf{Comparison of action recognition result under different 3D annotation quality.} We report PA-MPJPE (\si{\milli\meter}) of the estimated 3D hand-body poses compared with our multi-view annotation data, alongside action recognition accuracy (\%) on the AssemblyHands-X benchmark using MS-G3D~\cite{liu2020disentangling}.}
    \label{table:result}
\begin{adjustbox}{width=0.34\textwidth}
\begin{tabular}{@{}lllll@{}} \toprule
& & \multicolumn{2}{c}{PA-MPJPE (\si{\milli\meter})} &   \\ \cmidrule(lr){3-4} 
Annotation & \#view & All & Hand &Acc. \\ \midrule
SMPLer-X~\cite{cai2023smpler} & 1 & 43.78  & 15.43  & 59.7 \\
 + HaMeR~\cite{pavlakos2024reconstructing} & 1 & 45.47 & 10.62 & \underline{61.2} \\ \midrule
Ours & 8 & -  & - & \bfseries 69.6\\\bottomrule
\end{tabular}
\end{adjustbox}
\end{table}

Next, to investigate the impact of 3D pose annotation quality on action recognition, 
we evaluate both pose estimation using the latest single-view estimators and action recognition for varying inputs from these different pose estimators.
For single-view estimation, we use SMPLer-X~\cite{cai2023smpler}, a state-of-the-art 3D hand-body estimation model. 
Since SMPLer-X may produce noisy hand predictions, we also evaluate a variant where its hand estimates are replaced with those from HaMeR~\cite{pavlakos2024reconstructing}, a leading 3D hand estimator.
\\

\noindent\textbf{Results.} 
Table~\ref{table:result} reports the PA-MPJPE of the estimated 3D hand-body poses relative to our multi-view annotations, along with the resulting accuracy of a 3D pose-based action recognition model~\cite{liu2020disentangling}. 
Our multi-view annotations significantly outperform all single-view baselines (SMPLer-X and SMPLer-X + HaMeR) in action recognition accuracy,  demonstrating that higher-quality 3D poses contribute to better action recognition performance.

\section{Conclusion}
We introduce AssemblyHands-X, the first markerless 3D hand-body benchmark for bimanual activities, designed for the systematic evaluation of the effect of hand-body coordination in action recognition. Through extensive benchmarking, we demonstrate that modeling interdependent hand-body dynamics enhances a holistic understanding of bimanual activities.

\noindent\textbf{Acknowledgement.}
This work was supported by JSPS KAKENHI Grant Numbers JP23H00488 and JP24K02956.

{
    \small
    \bibliographystyle{ieeenat_fullname}
    \bibliography{main}

\begin{thebibliography}{22}
\providecommand{\natexlab}[1]{#1}
\providecommand{\url}[1]{\texttt{#1}}
\expandafter\ifx\csname urlstyle\endcsname\relax
  \providecommand{\doi}[1]{doi: #1}\else
  \providecommand{\doi}{doi: \begingroup \urlstyle{rm}\Url}\fi

\bibitem[Aganian et~al.(2023)Aganian, Stephan, Eisenbach, Stretz, and Gross]{aganian2023attach}
Dustin Aganian, Benedict Stephan, Markus Eisenbach, Corinna Stretz, and Horst-Michael Gross.
\newblock Attach dataset: Annotated two-handed assembly actions for human action understanding.
\newblock In \emph{Proceedings of the IEEE International Conference on Robotics and Automation (ICRA)}, pages 11367--11373, 2023.

\bibitem[Ben-Shabat et~al.(2021)Ben-Shabat, Yu, Saleh, Campbell, Rodriguez-Opazo, Li, and Gould]{ben2021ikea}
Yizhak Ben-Shabat, Xin Yu, Fatemeh Saleh, Dylan Campbell, Cristian Rodriguez-Opazo, Hongdong Li, and Stephen Gould.
\newblock The ikea asm dataset: Understanding people assembling furniture through actions, objects and pose.
\newblock In \emph{Proceedings of the IEEE/CVF Winter Conference on Applications of Computer Vision (WACV)}, pages 847--859, 2021.

\bibitem[Cai et~al.(2023)Cai, Yin, Zeng, Wei, Sun, Yanjun, Pang, Mei, Zhang, Zhang, et~al.]{cai2023smpler}
Zhongang Cai, Wanqi Yin, Ailing Zeng, Chen Wei, Qingping Sun, Wang Yanjun, Hui~En Pang, Haiyi Mei, Mingyuan Zhang, Lei Zhang, et~al.
\newblock Smpler-x: Scaling up expressive human pose and shape estimation.
\newblock In \emph{Proceedings of the Advances in Neural Information Processing Systems (NeurIPS)}, pages 11454--11468, 2023.

\bibitem[Comanici et~al.(2025)Comanici, Bieber, Schaekermann, Pasupat, Sachdeva, Dhillon, Blistein, Ram, Zhang, Rosen, et~al.]{comanici2025gemini}
Gheorghe Comanici, Eric Bieber, Mike Schaekermann, Ice Pasupat, Noveen Sachdeva, Inderjit Dhillon, Marcel Blistein, Ori Ram, Dan Zhang, Evan Rosen, et~al.
\newblock Gemini 2.5: Pushing the frontier with advanced reasoning, multimodality, long context, and next generation agentic capabilities.
\newblock \emph{arXiv preprint arXiv:2507.06261}, 2025.

\bibitem[Fan et~al.(2023)Fan, Taheri, Tzionas, Kocabas, Kaufmann, Black, and Hilliges]{fan2023arctic}
Zicong Fan, Omid Taheri, Dimitrios Tzionas, Muhammed Kocabas, Manuel Kaufmann, Michael~J Black, and Otmar Hilliges.
\newblock Arctic: A dataset for dexterous bimanual hand-object manipulation.
\newblock In \emph{Proceedings of the IEEE/CVF Conference on Computer Vision and Pattern Recognition (CVPR)}, pages 12943--12954, 2023.

\bibitem[Fan et~al.(2024)Fan, Ohkawa, Yang, Lin, Zhou, Zhou, Liang, Gao, Zhang, Zhang, et~al.]{fan2024benchmarks}
Zicong Fan, Takehiko Ohkawa, Linlin Yang, Nie Lin, Zhishan Zhou, Shihao Zhou, Jiajun Liang, Zhong Gao, Xuanyang Zhang, Xue Zhang, et~al.
\newblock Benchmarks and challenges in pose estimation for egocentric hand interactions with objects.
\newblock In \emph{Proceedings of the European Conference on Computer Vision (ECCV)}, pages 428--448, 2024.

\bibitem[Hartley and Zisserman(2003)]{hartley2003multiple}
Richard Hartley and Andrew Zisserman.
\newblock \emph{Multiple view geometry in computer vision}.
\newblock Cambridge university press, 2003.

\bibitem[Iskakov et~al.(2019)Iskakov, Burkov, Lempitsky, and Malkov]{iskakov2019learnable}
Karim Iskakov, Egor Burkov, Victor Lempitsky, and Yury Malkov.
\newblock Learnable triangulation of human pose.
\newblock In \emph{Proceedings of the IEEE/CVF International Conference on Computer Vision (ICCV)}, pages 7718--7727, 2019.

\bibitem[Kirillov et~al.(2023)Kirillov, Mintun, Ravi, Mao, Rolland, Gustafson, Xiao, Whitehead, Berg, Lo, et~al.]{kirillov2023segment}
Alexander Kirillov, Eric Mintun, Nikhila Ravi, Hanzi Mao, Chloe Rolland, Laura Gustafson, Tete Xiao, Spencer Whitehead, Alexander~C Berg, Wan-Yen Lo, et~al.
\newblock Segment anything.
\newblock In \emph{Proceedings of the IEEE/CVF International Conference on Computer Vision (ICCV)}, pages 4015--4026, 2023.

\bibitem[Lin et~al.(2019)Lin, Gan, and Han]{lin2019tsm}
Ji Lin, Chuang Gan, and Song Han.
\newblock Tsm: Temporal shift module for efficient video understanding.
\newblock In \emph{Proceedings of the IEEE/CVF International Conference on Computer Vision (ICCV)}, pages 7083--7093, 2019.

\bibitem[Liu et~al.(2019)Liu, Li, Chen, and Li]{liu2019soft}
Shichen Liu, Tianye Li, Weikai Chen, and Hao Li.
\newblock Soft rasterizer: A differentiable renderer for image-based 3d reasoning.
\newblock In \emph{Proceedings of the IEEE/CVF International Conference on Computer Vision (ICCV)}, pages 7708--7717, 2019.

\bibitem[Liu et~al.(2020)Liu, Zhang, Chen, Wang, and Ouyang]{liu2020disentangling}
Ziyu Liu, Hongwen Zhang, Zhenghao Chen, Zhiyong Wang, and Wanli Ouyang.
\newblock Disentangling and unifying graph convolutions for skeleton-based action recognition.
\newblock In \emph{Proceedings of the IEEE/CVF Conference on Computer Vision and Pattern Recognition (CVPR)}, pages 143--152, 2020.

\bibitem[Ohkawa et~al.(2023)Ohkawa, He, Sener, Hodan, Tran, and Keskin]{ohkawa2023assemblyhands}
Takehiko Ohkawa, Kun He, Fadime Sener, Tomas Hodan, Luan Tran, and Cem Keskin.
\newblock Assemblyhands: Towards egocentric activity understanding via 3d hand pose estimation.
\newblock In \emph{Proceedings of the IEEE/CVF Conference on Computer Vision and Pattern Recognition (CVPR)}, pages 12999--13008, 2023.

\bibitem[Pavlakos et~al.(2019)Pavlakos, Choutas, Ghorbani, Bolkart, Osman, Tzionas, and Black]{pavlakos2019expressive}
Georgios Pavlakos, Vasileios Choutas, Nima Ghorbani, Timo Bolkart, Ahmed~AA Osman, Dimitrios Tzionas, and Michael~J Black.
\newblock Expressive body capture: 3d hands, face, and body from a single image.
\newblock In \emph{Proceedings of the IEEE/CVF Conference on Computer Vision and Pattern Recognition (CVPR)}, pages 10975--10985, 2019.

\bibitem[Pavlakos et~al.(2024)Pavlakos, Shan, Radosavovic, Kanazawa, Fouhey, and Malik]{pavlakos2024reconstructing}
Georgios Pavlakos, Dandan Shan, Ilija Radosavovic, Angjoo Kanazawa, David Fouhey, and Jitendra Malik.
\newblock Reconstructing hands in {3D} with transformers.
\newblock In \emph{Proceedings of the IEEE/CVF Conference on Computer Vision and Pattern Recognition (CVPR)}, pages 9826--9836, 2024.

\bibitem[Rai et~al.(2021)Rai, Chen, Ji, Desai, Kozuka, Ishizaka, Adeli, and Niebles]{rai2021home}
Nishant Rai, Haofeng Chen, Jingwei Ji, Rishi Desai, Kazuki Kozuka, Shun Ishizaka, Ehsan Adeli, and Juan~Carlos Niebles.
\newblock Home action genome: Cooperative compositional action understanding.
\newblock In \emph{Proceedings of the IEEE/CVF Conference on Computer Vision and Pattern Recognition (CVPR)}, pages 11184--11193, 2021.

\bibitem[Sener et~al.(2022)Sener, Chatterjee, Shelepov, He, Singhania, Wang, and Yao]{sener2022assembly101}
Fadime Sener, Dibyadip Chatterjee, Daniel Shelepov, Kun He, Dipika Singhania, Robert Wang, and Angela Yao.
\newblock Assembly101: A large-scale multi-view video dataset for understanding procedural activities.
\newblock In \emph{Proceedings of the IEEE/CVF Conference on Computer Vision and Pattern Recognition (CVPR)}, pages 21096--21106, 2022.

\bibitem[Wu et~al.(2024)Wu, Zheng, Yang, Chen, Das, and Lu]{wu2024frequency}
Wenhan Wu, Ce Zheng, Zihao Yang, Chen Chen, Srijan Das, and Aidong Lu.
\newblock Frequency guidance matters: Skeletal action recognition by frequency-aware mixed transformer.
\newblock In \emph{Proceedings of the ACM International Conference on Multimedia (ACMMM)}, pages 4660--4669, 2024.

\bibitem[Yang et~al.(2023)Yang, Zeng, Yuan, and Li]{yang2023effective}
Zhendong Yang, Ailing Zeng, Chun Yuan, and Yu Li.
\newblock Effective whole-body pose estimation with two-stages distillation.
\newblock In \emph{Proceedings of the IEEE/CVF International Conference on Computer Vision (ICCV)}, pages 4210--4220, 2023.

\bibitem[Zhan et~al.(2024)Zhan, Yang, Zhao, Mao, Xu, Lin, Li, and Lu]{zhan2024oakink2}
Xinyu Zhan, Lixin Yang, Yifei Zhao, Kangrui Mao, Hanlin Xu, Zenan Lin, Kailin Li, and Cewu Lu.
\newblock Oakink2: A dataset of bimanual hands-object manipulation in complex task completion.
\newblock In \emph{Proceedings of the IEEE/CVF Conference on Computer Vision and Pattern Recognition (CVPR)}, pages 445--456, 2024.

\bibitem[Zheng et~al.(2023)Zheng, Lee, and Lu]{zheng2023ha}
Hao Zheng, Regina Lee, and Yuqian Lu.
\newblock Ha-vid: A human assembly video dataset for comprehensive assembly knowledge understanding.
\newblock In \emph{Proceedings of the Advances in Neural Information Processing Systems (NeurIPS)}, pages 67069--67081, 2023.

\bibitem[Zhou et~al.(2024)Zhou, Yan, Cheng, Yan, Dai, and Hua]{zhou2024blockgcn}
Yuxuan Zhou, Xudong Yan, Zhi-Qi Cheng, Yan Yan, Qi Dai, and Xian-Sheng Hua.
\newblock Blockgcn: Redefine topology awareness for skeleton-based action recognition.
\newblock In \emph{Proceedings of the IEEE/CVF Conference on Computer Vision and Pattern Recognition (CVPR)}, pages 2049--2058, 2024.

\end{thebibliography}
}

\end{document}